\documentclass[manuscript]{acmart}

\usepackage{graphicx}
\usepackage{multirow}
\usepackage{microtype}
\usepackage{tabularx}
\usepackage{booktabs}
\usepackage{xcolor}
\usepackage{caption}
\usepackage{subcaption}

\AtBeginDocument{%
  \providecommand\BibTeX{{%
    \normalfont B\kern-0.5em{\scshape i\kern-0.25em b}\kern-0.8em\TeX}}}

\begin{document}

\title{Dialoging Resonance: How Users Perceive, Reciprocate and React to Chatbot's Self-Disclosure in Conversational Recommendations}


\author{Kai-Hui Liang}
\email{kaihui.liang@columbia.edu}
\affiliation{%
  \institution{Columbia University}
  \city{New York}
  \state{NY}
  \country{USA}
  \postcode{10027}
}
\author{Weiyan Shi}
\email{ws2634@columbia.edu}
\affiliation{%
  \institution{Columbia University}
  \city{New York}
  \state{NY}
  \country{USA}
  \postcode{10027}
}
\author{Yoojung Oh}
\email{yjeoh@ucdavis.edu}
\affiliation{%
  \institution{University of California, Davis}
  \city{Davis}
  \state{CA}
  \country{USA}
  \postcode{95616}
}

\author{Hao-Chuan Wang}
\email{hciwang@ucdavis.edu}
\affiliation{%
  \institution{University of California, Davis}
  \city{Davis}
  \state{CA}
  \country{USA}
  \postcode{95616}
}

\author{Jingwen Zhang}
\email{jwzzhang@ucdavis.edu}
\affiliation{%
  \institution{University of California, Davis}
  \city{Davis}
  \state{CA}
  \country{USA}
  \postcode{95616}
}
\author{Zhou Yu}
\email{zy2461@columbia.edu}
\affiliation{%
  \institution{Columbia University}
  \city{New York}
  \state{NY}
  \country{USA}
  \postcode{10027}
}

\renewcommand{\shortauthors}{Liang et al.}

\begin{abstract}
Using chatbots to deliver recommendations is increasingly popular. The design of recommendation chatbots has primarily been taking an information-centric approach by focusing on the recommended content per se. Limited attention is on how social connection and relational strategies, such as self-disclosure from a chatbot, may influence users' perception and acceptance of the recommendation. In this work, we designed, implemented, and evaluated a social chatbot capable of performing three different levels of self-disclosure: factual information (low), cognitive opinions (medium), and emotions (high). In the evaluation, we recruited 372 participants to converse with the chatbot on two topics: movies and COVID-19 experiences. In each topic, the chatbot performed small talks and made recommendations relevant to the topic. Participants were randomly assigned to four experimental conditions where the chatbot used factual, cognitive, emotional, and adaptive strategies to perform self-disclosures. By training a text classifier to identify users' level of self-disclosure in real-time, the adaptive chatbot can dynamically match its self-disclosure to the level of disclosure exhibited by the users. Our results show that users reciprocate with higher-level self-disclosure when a recommendation chatbot consistently displays emotions throughout the conversation. Chatbot's emotional disclosure also led to increased interactional enjoyment and more positive interpersonal perception towards the bot, fostering a stronger human-chatbot relationship and thus leading to increased recommendation effectiveness, including a higher tendency to accept the recommendation. We discuss the understandings obtained and implications to future design.



\end{abstract}



\maketitle

\section{Introduction}
  
As chatbots are becoming increasingly popular across various digital platforms, chatbot-human conversations have evolved into more complex and extended exchanges that often invoke disclosing personal information and emotional experiences. In recent years, developments in natural language processing have empowered chatbots with a greater capacity to engage in social and emotional conversations. In addition, endeavors have been made to create and test persuasive or recommendation chatbots that initiate recommendations to users to adopt. A study showed that people preferred a social chatbot that could remember their preferences and make valuable recommendations \cite{thies2017you}. This represents the current approach in developing persuasive chatbots, that is, exploring more effective ways in eliciting user preferences \cite{ikemoto2019tuning, kim2020study, radlinski2019coached, gao2021advances} and optimizing the presentation of recommendations. For instance, \cite{ikemoto2019tuning} sought to optimize the order and numbers of the preference questions being asked before making a recommendation. When presenting recommendations, many existing chatbots focus on showing item information (e.g., movie plot summary, song artist, a link), and others further provide explanations of the recommendations. For example, \cite{wilkinson2021or} demonstrated that justifying why an item is selected in a recommendation chatbot enhances users’ trust and use experience. 

However, these studies so far have only emphasized the recommended content itself by providing more information and explanation about it. This information-centric approach neglects and underutilizes the chatbot's characteristics relating to social identity and human-chatbot relationship. As many chatbots are personified and equipped with social conversation capabilities, there have been few investigations concerning how social connection and relational strategies would influence persuasion and recommendation effectiveness. In this work, we aim to expand this research direction and evaluate the effects of chatbot's self-disclosures on users' perceptions of the chatbot and given recommendations. By emphasizing on self-disclosure, we highlight the importance of considering the role of chatbot's social identity and relationship strategies in human-chatbot conversations.    

Self-disclosure is a fundamental aspect of human communication. It is the act of disclosing personal information to others, such as personal experiences, opinions, thoughts, beliefs, and feelings \cite{barak2007degree}. Prior studies categorized people's self-disclosure into three levels, assuming that the three levels involve different depths of disclosures. The first level is a peripheral level of disclosure, involving disclosing one's basic information. It is also conceptualized as factual disclosure. The second level is an intermediate level of disclosure, involving disclosing one's thoughts and preferences, or cognitive assessments of the information. The third, or core level, involves disclosures of one's emotions, or emotional experiences concerning a situation or object\cite{malloch2019seeing, altman1973social, barak2007degree, lee2020hear}. Hierarchically, factual, cognitive, and emotional disclosures represent three different depths in disclosure. As people move from factual to emotional, the expectation is that deeper senses of social relationships will be built. 

Applications of self-disclosures to the design of recommendation chatbots have been initiated in the past. \cite{hayati2020inspired} revealed that recommendation strategies related to self-disclosure such as conveying a recommender's ``personal opinion'' and demonstrating ``similarity'' between the recommender and a seeker's views have a significant positive effect on users' acceptance of the recommendations among human-human conversations. It has also been shown that higher self-disclosure from a chatbot leads to users' more positive perception of the chatbot (more trust and interactional enjoyment) in a conversational movie recommendation system \cite{lee2017enhancing}. However, these studies only showed the effects of chatbot's disclosures on users' perceptions of the chatbot, without clear suggestions on how users evaluate the recommendations or their willingness to accept the recommendations. In addition, no work has systematically assessed how chatbots utilizing different levels of self-disclosures may impact their interactions with the users differently.

In this work, following the three-level self-disclosure framework from the literature of human communication \cite{malloch2019seeing, altman1973social, barak2007degree, lee2020hear}, we designed a social chatbot with three levels of self-disclosures: factual (low), cognitive (medium), and emotional (high). We recruited 372 participants to converse with the chatbot on two topics: movies, and COVID-19. In each topic session, the chatbot performed small talks and made recommendations relevant to the topic, such as recommending a particular movie or suggesting self-care practices to relieve anxiety. Participants were randomly assigned to four experiment groups where the chatbot used (1) factual, (2) cognitive, (3) emotional, and (4) adaptive self-disclosures. The fourth group had the chatbot dynamically matching the user's self-disclosure strategies. Results showed that emotional self-disclosure was more effective as users adopted a more positive attitude towards chatbot's recommendations. Moreover, the chatbot with emotional disclosure elicited more positive interpersonal perception towards the chatbot, such that participants engaged more in the conversation and perceived the chatbot as warmer (e.g., more friendly, kind). Lastly, we observed evidence suggesting that people reciprocated the chatbot's self-disclosure levels. These imply that when designing persuasive or recommendation chatbots, emotional self-disclosures can be incorporated as a conversational strategy to enhance the human-chatbot relationship and lead to more effective conversations that render more successful in recommendation acceptance. Further, it suggests chatbot can to some extent lead the conversation and use self-disclosures to encourage more human disclosures.

Our work makes three major contributions to the communities. First, we apply the self-disclosure theory in building more effective human-chatbot conversations and demonstrate the utility of involving emotional disclosures in chatbot conversations. Second, we empirically differentiated the effects of chatbot's factual, cognitive, and emotional disclosure's impact on human perceptions of chatbot and the recommendations made by the chatbot. Last, we used two different topics, movie recommendation and COVID discussions, demonstrating the generalizability of the contribution of emotional self-disclosure. Given the potential for chatbots to play a more prominent role in human-machine communication and in supporting human decisions, our research is of value in guiding chatbot-based conversational designs. Ethical considerations are discussed in the end of the paper.

\section{Backgrounds} 
 
\subsection{Levels of Self-disclosure}
Self-disclosure is the act of disclosing personal information to others in interpersonal communication contexts. According to social penetration theory (SPT) \cite{altman1973social}, self-disclosure is the intentional revealing of personal information such as personal opinions, thoughts, beliefs, feelings, and experiences \cite{altman1973social}. Self-disclosure helps interpersonal relationships progress, and this proposition of the SPT essentially holds true in computer-mediated communication \cite{altman1973social, hu2004friendships, huang2016examining}. Self-disclosure can be characterized by the levels of depth in disclosure. Previous research has proposed and employed a framework that categorized people's self-disclosure into three levels: peripheral disclosure of one's information (i.e., factual), intermediate disclosure of one's thoughts (i.e., cognitive), and core-level disclosure of one's emotions (i.e., emotional) \cite{malloch2019seeing, altman1973social, barak2007degree, lee2020hear, malloch2019seeing}. Our study follows this framework and designs our chatbot with the three self-disclosure levels.

\subsection{The Effect of Self-disclosure on Reciprocity}

Reciprocity is an indispensable component in social penetration theory and has been shown to be one of the most significant outcomes of self-disclosure \cite{jourard1971self}. Previous studies have demonstrated that self-disclosure induces reciprocity \cite{dindia2002self, barak2007degree}. That is, as one party discloses themselves, the other party would be more inclined to self-disclose. Such an effect of self-disclosure on reciprocity can vary by the level of disclosure. For example, in Barak and Gluck-ofri \cite{barak2007degree}, they examined the reciprocity of self-disclosure by three different levels of self-disclosure (i.e., information, thoughts, feelings) in online forums. Their work measured the reciprocity of self-disclosure by calculating the correlation between the self-disclosure of people's postings and the self-disclosure expressed in reacting to these postings. Results showed that significant reciprocity was found in all levels of self-disclosure (information, thoughts, and feelings). Among the disclosure types, emotional reciprocity was the strongest. 

In human-chatbot conversation, initial evidence has shown that people reciprocate to chatbots' self-disclosure. For example, Ravichander et al. \cite{ravichander2018empirical} demonstrated that when the bot self-discloses, the user self-discloses in response. Lee et al. \cite{lee2020hear} also shared their findings that a high self-disclosing chatbot (i.e., deeper feelings, thoughts, or facts) induces people to reveal deeper thoughts and more feelings on sensitive topics than a low self-disclosing one (i.e. fewer feelings and thoughts.) However, how the effect of chatbots' emotional disclosure differs from factual or cognitive disclosures remains unknown. In this work, we explore how people reciprocate to chatbots' different self-disclosure levels in the context of chatbot-based recommendations. Specifically, the chatbot's disclosure is designed into finer-grained levels (factual, cognitive, and emotional), and the users' self-disclosure levels are measured with the same three levels to evaluate reciprocity.

\subsection{The effect of Self-disclosure on Perception of a Chatbot}
Previous studies have demonstrated that self-disclosure is related to positive interpersonal perceptions of the conversational partner such as liking and intimacy \cite{collins1994self, dindia2002self, berg1987themes, lee2017enhancing}. Furthermore, evidence shows that self-disclosure can also elicit positive perceptions (i.e., intimacy) towards a non-human agent such as a computer and chatbots. For instance, Moon found that people develop intimacy with computers through self-disclosure and reciprocity \cite{moon2000intimate}. Similarly, studies found that chatbot's self-disclosure had a positive effect on increasing participants' perceived intimacy towards the bot \cite{lee2020hear} and build trust with the bot \cite{lee2017enhancing}. Furthermore, different levels of self-disclosure can induce interpersonal perceptions differently. Ho et al. \cite{ho2018psychological} manipulated the levels of self-disclosure into factual (i.e., objective information) self-disclosure and emotional (i.e., expression of emotions and feelings) self-disclosure and found that the effects of emotional disclosure were more substantial than factual disclosure, especially on perceptions of partner's warmth. In the current study, we design our chatbot with finer-grained levels of self-disclosure (i.e., factual, cognitive, emotional, and adaptive) and investigate how different levels would influence an individual's perceptions of the chatbot. 


\subsection{The effect of Self-disclosure on Relationship and Recommendation Success}
Self-disclosure is particularly important because it contributes to building relationships between humans and chatbots. According to SPT, self-disclosure helps interpersonal relationship progress (\cite{altman1973social}). As the speaker increases his/her disclosure level and opens up his/her mind, the listener feels closer to the speaker {\cite{laurenceau1998intimacy, reis1988intimacy, han2011empathic}}. Previous studies on relational agents have utilized self-disclosure to develop and maintain positive relationships between humans and bots. They have shown that such relationships can enhance the effectiveness of the bot. It is suggested that chatbots' self-disclosure builds trust with users and makes the conversation more enjoyable {\cite{ho2018psychological, lee2017enhancing}} which is vital to understanding whether users would continue to use the system. For instance, in Bickmore et al. \cite{bickmore2013tinker}, a museum guide agent used reciprocal self-disclosure to establish social bonds with users, and this led to continued interaction with the chatbot and repeated visits to the museum. In an example of a movie recommendation system, chatbot's self-disclosure increased users' intention to use the system \cite{lee2017enhancing}. 

However, the limitation of such a perspective is that measuring users' intention to use the chatbot may not represent the effectiveness of the delivered content (e.g., recommendations). Thus, it is necessary to gauge whether the delivered content would potentially change users' attitudes or intentions towards chatbot's recommendations. Initial evidence of human-human conversations on movie \cite{hayati2020inspired} revealed that a recommender's disclosure of personal opinions and similarity to the user had a positive effect on users' acceptance of the recommendation. Yet, the study did not separate out the emotional element within the self-disclosure strategy, making it difficult to measure the effects of emotional self-disclosure. To the best of our knowledge, no study has attempted to investigate the effects of chatbot's self-disclosure on the recommendation's persuasiveness (e.g., changes in attitudes and/or intentions to follow).

Overall, in our study, we explore the following three research questions: 
\textbf{RQ1}: \textit{How do people reciprocate to different levels of self-disclosure (i.e., factual, cognitive, emotional)} performed by a chatbot? 
\textbf{RQ2}: \textit{Can different levels of self-disclosure (i.e., factual, cognitive, emotional) of a chatbot influence people's conversation engagement and perception of the bot?} 
\textbf{RQ3}: \textit{Can different levels of self-disclosure (i.e., factual, cognitive, emotional) of a chatbot impact people's enjoyment of receiving chatbots' recommendations, attitudes towards the recommendation, and intention to follow it?}

\section{Chatbot Design}
To answer our research questions, we designed and implemented a text-based chatbot capable of different levels of self-disclosure. We will first introduce the chatbot's architecture. Then we talk about the two topics of the conversations. Finally, we cover different stages of the conversations, small talks and recommendations, along with their dialog template designs. 

\subsection{Chatbot System Architecture}
We built the text-based chatbot system using the Amazon Conversational Bot Toolkit (Cobot) \cite{khatri2018advancing}. Each conversational turn first goes to the AWS Lambda function as a RESTful API event request through the Amazon API gateway. Then the information passes to a natural language understanding module, which borrowed from Gunrock 2.0 (an Alexa Prize Socialbot) \cite{liang2020gunrock}. This module contains critical components such as sentence segmentation, dialog act prediction, movie name entity recognition. The understanding model then outputs information to the dialog management module, which is adapted from Gunrock 2.0 as well. This management module stores participant attributes and dialog states in DynamoDB, and uses a custom Finite State Machine manager to handle dialog state transition. Finally the information passes down to the response generation module.

The response generation module consists of a dynamic response generator and a topic proposer. The dynamic response generator is made of a question handler to answer users' questions, and an acknowledgment generator to react to users' other responses. (More details are described in Appendix \ref{subsubsec:acknowledgement_generator} and  \ref{subsubsec:question_handler}.) Inspired by Gunrock 2.0 \cite{liang2020gunrock}, the dynamic response generator is designed to generate plausible, meaningful, and natural responses. Note that it leverages a conversational generation model, Blender \cite{roller2020recipes}, which is trained on large amounts of human-human conversation data and is adjusted to learn several conversational skills from human. By incorporating the generation model, we provide engaging responses that can adapt to a wide range of user responses. The topic proposer is made of predefined templates designed for different self-disclosure levels (\ref{sec:self-disclosure-design}). At every dialog turn, the chatbot first replies with the dynamic response generator, and concatenates the response with the template selected by the topic proposer.


\subsection{Dialog Sessions}

\begin{figure}[t]%
\centering
\includegraphics[width=1.0 \linewidth]{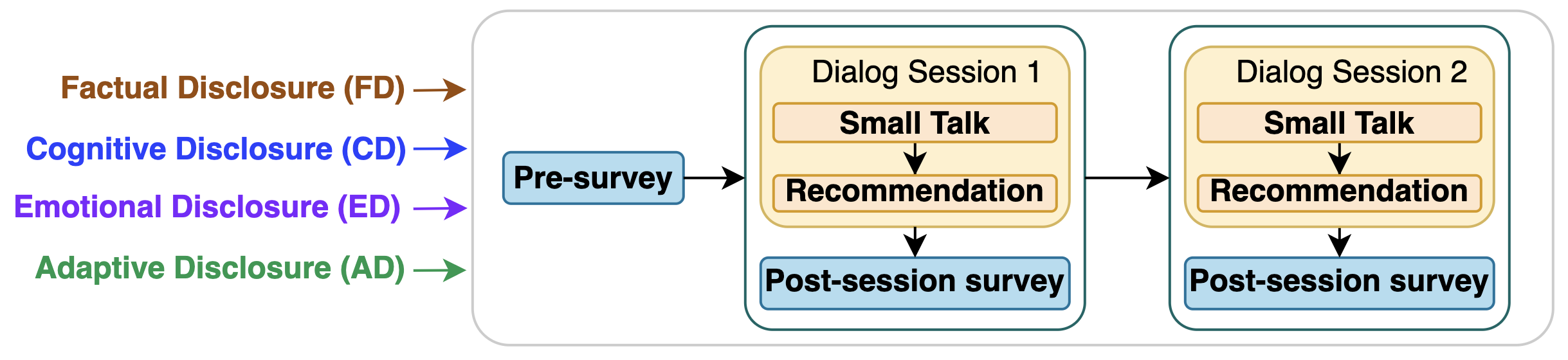}
    \caption{Study procedure: First, we randomly assigned the participants to one of four self-disclosure groups: factual disclosure (FD), cognitive disclosure (CD), emotional disclosure (ED), and adaptive disclosure (AD). After filling in a pre-survey, participants engaged in two dialog sessions and filled in the same post-session survey after each session. }
    \vspace{-3mm} 
    \label{fig:procedure}
\end{figure}

\label{sec:dialog_sessions}
As illustrated in Fig. \ref{fig:procedure}, we asked participants to engage in two dialog sessions with the chatbot. In each session, the bot talked about one topic, either Movie or COVID-19 experiences. For each topic, the bot first performed \textit{Small-talk} (\ref{subsubsec:small_talk}) with the users, which was the treatment of this study, and then it voluntarily provided a \textit{Recommendation} (\ref{subsubsec:recommendation}) relevant to the current topic to the users. Afterward, the chatbot wrapped up the session. 

We used Movie and COVID-19 for topics, because most participants relate to them and share their own ideas. We selected these two intrinsically different topics to mitigate topic bias. Watching movies is a common entertaining activity for people. We expect our results to be applicable to other product recommendations. COVID-19 is a global pandemic, which is not only applicable to the majority of people but also relates to their well-being. Given that more recent works investigate building chatbots for emotional and behavior-change support during the pandemic by letting participants disclose their situation and concerns or by suggesting self-care tips \cite{miner2020chatbots}, our study may be beneficial in these contexts.

To evaluate the effects of different levels of self-disclosure exhibited by the chabot, we randomly assigned participants to one of the four experimental groups: \textbf{Factual (FD), Cognitive (CD), Emotional (ED), and Adaptive (AD) self-disclosure groups}. In the \textit{Small Talk} phase, the highest level of disclosure that participants would experience was based on the experimental condition they were assigned to.
Many studies have indicated that the level of self-disclosure is positively related to interlocutors' mutual relationship and intimacy \cite{west2013introducing, malloch2019seeing, altman1973social, barak2007degree, lee2020hear, malloch2019seeing}. We consider the progress of the human-chatbot relationship stable within the study period (two dialog sessions within about 30 minutes). Hence, the highest level of disclosure exhibited by the chatbot throughout the study could serve as a social cue, hinting to users the extent of affinity it is "willing" to have with the listener. 

To ensure the salience of a chabot's self-disclosure, we tuned the chatbot to exhibit the highest self-disclose level, matching the experimental condition, at each conversation turn. However, when it is unnatural for the chatbot to self-disclose, such as when the chatbot needs to confirm information received (e.g., the movie name), exceptions were made not to exhibit self-disclosure. Besides, it is considered unnatural to stick to only a single level of disclosure for the whole conversation (e.g., being emotional throughout the discourse). Instead, a more natural conversation may arguably have a mixture of different types of disclosure once in a while. For example, \textit{``I don't like horror movies because they are disgusting [cognitive] and I'm too scared to watch them [emotional]''} includes both cognitive and emotional disclosure at the same dialog turn. Therefore, the levels of self-disclosure are incremental from one condition to another. In FD, the disclosure levels available for the chatbot to exhibit in a turn are none or factual-level disclosure; In CD, the chatbot can exhibit none or factual- or cognitive-level disclosure; and in ED, a mixture of all levels of disclosure are possible.

\begin{figure*}[htb]%
\centering
\includegraphics[width=1.0\linewidth]{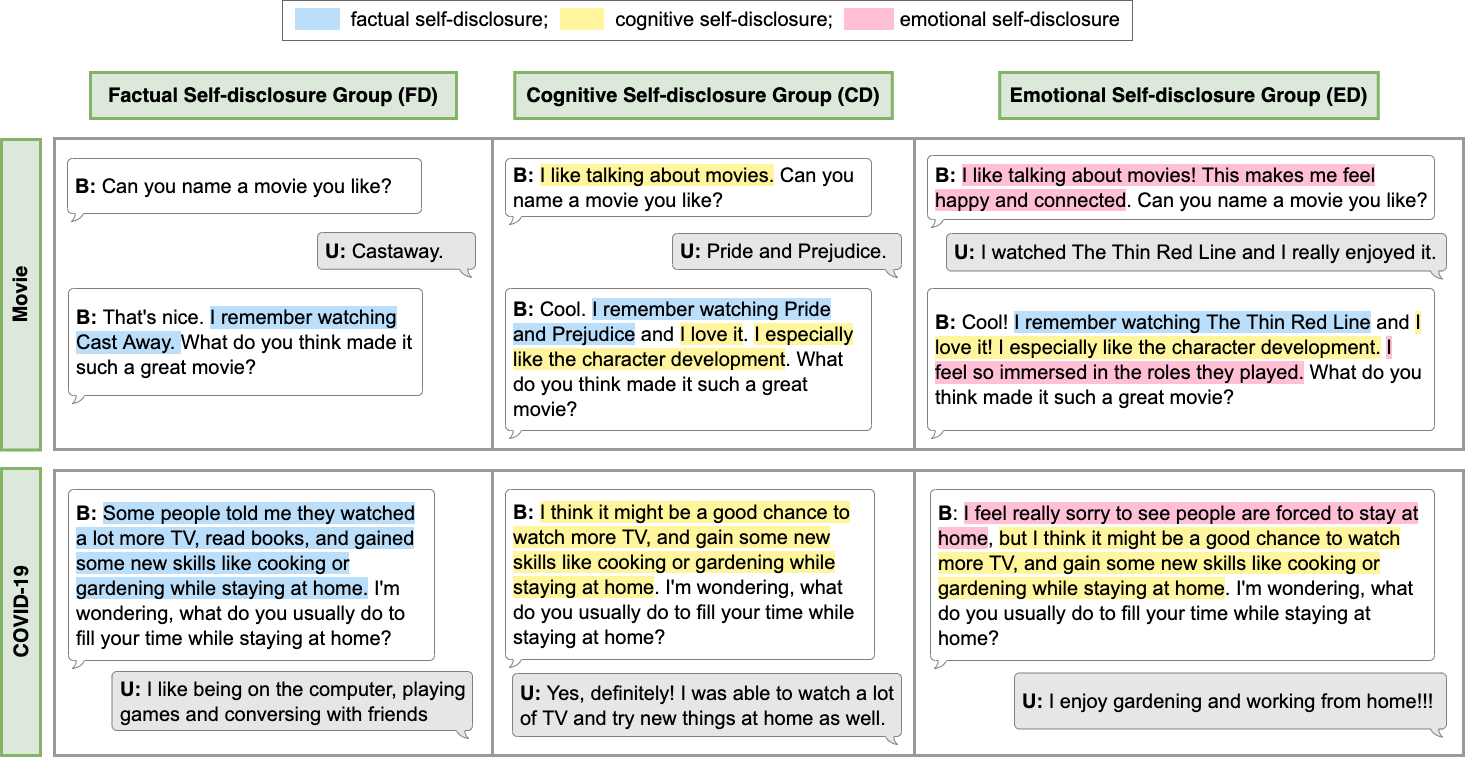}
    \caption{Small talk dialog excerpts of movie and COVID-19 sessions in the factual disclosure (FD), cognitive disclosure (CD), and emotional disclosure (ED) groups.}
    \vspace{-3mm} 
    \label{fig:small_talk_dialog_excerpt}
\end{figure*}


\subsubsection{Small Talk}
\label{subsubsec:small_talk}
\paragraph{Subtopics} The small talks in both movie and COVID-19 sessions contain six to seven subtopics. (See Fig. \ref{fig:small_talk_dialog_excerpt} for dialog excerpts. Full dialog example can be found in Fig. \ref{fig:dialog_example_movie} and \ref{fig:dialog_example_covid}.)
The subtopics are the same across all groups, while the specific bot utterances differ depending on the corresponding self-disclosure level. The bot proposes subtopics turn by turn. In each turn, the bot first replies to the participant's utterance to ensure the bot is responsive to the participants. Then the bot proposes the next subtopic in the same turn. In the \textit{Movie Session}, the bot first asks a movie the users like (see Appendix \ref{sec:movie_grounding} for movie recognition details), and continues the discussion by asking users' opinions and providing interesting facts. It then asks users' favorite actors and movie genre preferences. In \textit{COVID-19 Session}, the bot discusses participants' experiences during the pandemic, such as activities they did during the quarantine, their opinions toward social distancing, and their changes in shopping and diet. 



\paragraph{Self-disclosure template design} 
\label{sec:self-disclosure-design} 
For each subtopic, we design different response templates for each self-disclosure level. Figure \ref{fig:small_talk_dialog_excerpt} demonstrates small talk dialog excerpts for both topics in all groups.
In FD, the bot at most provides facts about itself (e.g., \textit{``I remember watching Titanic''}) and does not share its opinions, preferences, or emotion. If the bot asks opinion or preference questions to the user (e.g., \textit{``What parts of this movie did you like the best? ''}), it does so without telling its own. In CD, in addition to facts, it shares its opinion (e.g., how it thinks of social distancing) without revealing its emotion. In ED, the bot further shares its feelings or expresses empathy along with facts and opinions. In AD, the bot starts with the factual level disclosure in the beginning of the conversation. At every turn, it detects the user's disclosure level and matches its disclosure level to the user's highest self-disclosure level ever. That is, the bot adapts its highest self-disclosure level reciprocally to the user's highest level. We design it this way given that users' highest level of disclosure reflects the extent to which the user is willing to reveal to the bot, and the bot discloses to a higher level only if the user takes the initiative.

\subsubsection{Recommendation}
\label{subsubsec:recommendation}
To investigate whether the bot's different self-disclosure levels impact the recommendation's effectiveness (RQ3), the bot voluntarily gives relevant recommendations at the end of each session (\ref{sec:movie_recommendation_rule}). In the movie session, the bot recommends a movie to the participants depending on the participants' preferences collected in the \textit{Small Talk}. In the COVID-19 session, the bot suggests self-care practices such as unplugging from technology and taking a walk. More details and dialog examples can be found in \ref{sec:full_dialog_design} and \ref{sec:movie_recommendation_rule}.

\section{Study Method}

\subsection{Recruitment and Participants}
We recruited participants on Amazon Mechanical Turk. To ensure the participants had something to say to our chatbot, we recruited participants who self-identified to be movie lovers. We filtered out conversations where participants did not finish the tasks, or encountered system errors. We in total have 372 participants. We randomly assign users to one of the four conditions of the system.
There are 106 participants in the factual disclosure condition, 91 in the cognitive disclosure condition, 85 in the emotional disclosure condition, and 90 in the adaptive disclosure condition. We deployed our bot on a web interface where the participants could chat with the bot in typing text.

\subsection{Procedure}
Figure \ref{fig:procedure} shows the user study procedure. The participants were first randomly assigned to one of four self-disclosure groups. Then, they filled out a pre-task survey that collected demographic information. Afterward, they began the first dialog session on the randomly assigned topic (which could be movie or COVID-19) and filled out a post-session survey (\ref{sec:survey}). After that, they started the second dialog session, where they discussed the other topic and filled in the same post-session survey as the previous session.  



\subsection{Measurement}
The quantitative measures were obtained from conversational dialogs and self-reported surveys.

\subsubsection{Conversation Log Analysis} 
We utilized LIWC2015 \cite{tausczik2010psychological} to calculate the \textit{word length}. Previous work showed that word count is positively related to self-disclosure \cite{kreiner2019self}. In addition, to detect participants' self-disclosure level that aligns with our scheme, we developed a self-disclosure classifier to detect participants' self-disclosure level in each turn (Sec. \ref{sec:self-disclosure-clssifier}).

\subsubsection{Post-session Survey}
\label{sec:survey}
After each dialog session, we collected user self-reported ratings on two dimensions: user engagement, the participant's perception of the bot (five constructs), and the bot's recommendation effectiveness (three constructs). To ensure the robustness of the constructs, we used three measurement items with 5-point Likert scales for each construct and calculated the average score to represent the construct's value. The measurement items were adapted from prior literature and modified to fit the context of human-chatbot interaction. Table \ref{fig:survey_measurement_perception} shows the measurement items of each construct for engagement and perception of the bot, and Table \ref{fig:survey_measurement_recommendation} shows the recommendation effectiveness. All constructs' alpha values are higher than 0.80, indicating satisfactory reliabilities. We also asked two open-ended questions in the survey to collect some qualitative opinions from the participants.

\paragraph{Engagement and perceptions of the bot.} 
We measured ``Engagement'' to see how much people enjoy the conversation, which is an essential indication of people's willingness to continue the conversation. ``Closeness'' was measured using three items we developed as a close relationship is often built by self-disclosure behavior. We measured ``Warmth'' to see how friendly/sympathetic/kind the participants think of the bot. We also measured ``competence'' to see how participants consider the bot's ability to conduct a conversation; ``Humanlikeness'' to understand how much they perceived the bot as humans; ``Eeriness''  to see if they think the bot is weird.

\paragraph{Recommendation effectiveness.} ``Recommendation attitude'' measured the extent to which participants \textit{cognitively} agree with the recommendation. We also measured ``recommendation enjoyment'' to understand if the participants \textit{emotionally} enjoy listening to the recommendation. Even if people enjoy and have positive attitudes towards the recommendation, that does not necessarily mean they plan to take action, so we measured ``recommendation intetion'' to see how much people intend to follow the recommendation.

\paragraph{Opinion questions.}
To probe participants' opinions on the conversation, we asked two open questions at the end of the survey: \textit{``Which part of the conversation did you like best?''} and  \textit{``Which part of the conversation did you like least?''}


\begin{table}[t!]
\fontsize{8}{10}\selectfont

\begin{tabular}{p{1.7cm}p{4.5cm}p{3.5cm}cc}
\hline
\textbf{Construct}                                    & \multicolumn{1}{c}{\textbf{Measurement item}}                  & \multicolumn{1}{c}{\textbf{Scale}} & \textbf{\begin{tabular}[c]{@{}c@{}}Cronbach's \\ alpha\end{tabular}} & \textbf{Ref.} \\ \hline
\multirow{3}{*}{\textbf{Engagement}}                  & \textit{How engaging did you feel during the conversation?}    & 1=unappealing, 5=engaging          & \multirow{3}{*}{0.907}                                               & \multirow{3}{*}{\cite{hassanein2007manipulating, ghani1994task}}  \\
                                                      & \textit{How enjoyable did you feel during the conversation?}   & 1=unpleasant, 5=enjoyable          &                                                                      &                    \\
                                                      & \textit{How interesting did you feel during the conversation?} & 1=boring, 5=interesting            &                                                                      &                    \\ \hline
\multirow{3}{2cm}{\textbf{Perceived closeness}}         & \textit{How close did you feel with the bot?}                  & 1=distant, 5=close                 & \multirow{3}{*}{0.946}                                               & \multirow{3}{*}{}  \\
                                                      & \textit{How connected did you feel with the bot?}              & 1=unconnected, 5=connected         &                                                                      &                    \\
                                                      & \textit{How associated did you feel with the bot?}             & 1=disassociated, 5=associated      &                                                                      &                    \\ \hline
\multirow{3}{2cm}{\textbf{Perceived bot warmth}}        & \textit{How friendly did you find the bot?}                    & 1=distant, 5=friendly              & \multirow{3}{*}{0.808}                                               & \multirow{3}{*}{\cite{bartneck2009measurement, wiggins1988psychometric}}  \\
                                                      & \textit{How sympathetic did you find the bot?}                 & 1=unsympathetic, 5=sympathetic     &                                                                      &                    \\
                                                      & \textit{How kind did you find the bot?}                        & 1=cold-hearted, 5=kind             &                                                                      &                    \\ \hline
\multirow{3}{2cm}{\textbf{Perceived bot competence}}    & \textit{How coherent did you feel during the conversation?}    & 1=incoherent, 5=coherent           & \multirow{3}{*}{0.864}                                               & \multirow{3}{*}{}  \\
                                                      & \textit{How rational did you feel during the conversation?}    & 1=irrational, 5=rational           &                                                                      &                    \\
                                                      & \textit{How reasonable did you feel during the conversation?}  & 1=reasonable, 5=unreasonable       &                                                                      &                    \\ \hline
\multirow{3}{2cm}{\textbf{Perceived bot humanlikeness}} & \textit{How human-like did you find the bot?}                  & 1=fake, 5=human-like               & \multirow{3}{*}{0.932}                                               & \multirow{3}{*}{\cite{bartneck2009my}}  \\
                                                      & \textit{How natural did you find the bot?}                     & 1=machine-like, 5=lifelike         &                                                                      &                    \\
                                                      & \textit{How lifelike did you find the bot?}                    & 1=artificial, 5=enjoyable          &                                                                      &                    \\ \hline
\multirow{3}{2cm}{\textbf{Perceived bot eeriness}}      & \textit{How weird did you find the bot?}                       & 1=normal, 5=weird                  & \multirow{3}{*}{0.890}                                               & \multirow{3}{*}{\cite{gray2012feeling, ho2017measuring}}  \\
                                                      & \textit{How creepy did you find the bot?}                      & 1=pleasant, 5=creepy               &                                                                      &                    \\
                                                      & \textit{How freaked out did you find the bot}                  & 1=ordinary, 5=freaked out          &                                                                      &                    \\ \hline
\end{tabular}
    \caption{Constructs and measurement items for engagement and people's perception of the bot.}
    \label{fig:survey_measurement_perception}
\end{table}

\begin{table}[t!]
\fontsize{8}{10}\selectfont
\begin{tabular}{p{1.3cm}p{5.9cm}lcl}
\hline
\textbf{Construct}                  & \textbf{Measurement item}                                       & \textbf{Scale}                                                                                      & \textbf{\begin{tabular}[c]{@{}c@{}}Cronbach's \\ alpha\end{tabular}} & \textbf{Ref.} \\ \hline
\multirow{3}{*}{\textbf{Enjoyment}} & I like the bot's recommendation at the end.                     & \multirow{3}{*}{\begin{tabular}[c]{@{}l@{}}1=strongly disagree, \\ 5 = strongly agree\end{tabular}} & \multirow{3}{*}{0.937}                                               & \multirow{3}{*}{\cite{hassanein2007manipulating, ghani1994task}}  \\
                                    & I enjoy hearing the bot's recommendation at the end.            &                                                                                                     &                                                                      &                    \\
                                    & The bot's recommendation at the end sounds interesting to me.   &                                                                                                     &                                                                      &                    \\ \hline
\multirow{3}{*}{\textbf{Attitude}}  & It is useful to follow the bot's recommendation.                & \multirow{3}{*}{\begin{tabular}[c]{@{}l@{}}1=strongly disagree, \\ 5 = strongly agree\end{tabular}} & \multirow{3}{*}{0.947}                                               & \multirow{3}{*}{\cite{ajzen2006constructing}}  \\
                                    & It is wise to follow the bot's recommendation.                  &                                                                                                     &                                                                      &                    \\
                                    & It is beneficial to follow the bot's recommendation at the end. &                                                                                                     &                                                                      &                    \\ \hline
\multirow{3}{*}{\textbf{Intention}}  & I plan to follow the bot's recommendation.                      & \multirow{3}{*}{\begin{tabular}[c]{@{}l@{}}1=strongly disagree, \\ 5 = strongly agree\end{tabular}} & \multirow{3}{*}{0.968}                                               & \multirow{3}{*}{\cite{fishbein2011predicting, rhodes2005discrepancies}}  \\
                                    & I intend to follow the bot's recommendation.                    &                                                                                                     &                                                                      &                    \\
                                    & It is my intention to follow the bot's recommendation.          &                                                                                                     &                                                                      &                    \\ \hline
\end{tabular}
    \caption{Constructs and measurement items for recommendation effectiveness.}
    \label{fig:survey_measurement_recommendation}
\end{table}






\section{Self-disclosure Level Classifier} 
\label{sec:self-disclosure-clssifier}
To build a chatbot capable of adapting its self-disclosure to users' level during the conversations (the AD condition), we need to identify users' self-disclosure levels in real-time. Therefore, we created a self-disclosure level classifier with a BERT-based model \cite{devlin2018bert}, and fine-tuned it on our own annotated dataset. As the classifier was proven reliable, the same classifier was also used in post-hoc analysis to evaluate users' reciprocity by identifying self-disclosure levels from the users in the FD, CD, and ED groups.

\subsection{Annotation Scheme} 
\begin{figure*}[ht!]
    \centering
    \includegraphics[width=0.82\linewidth]{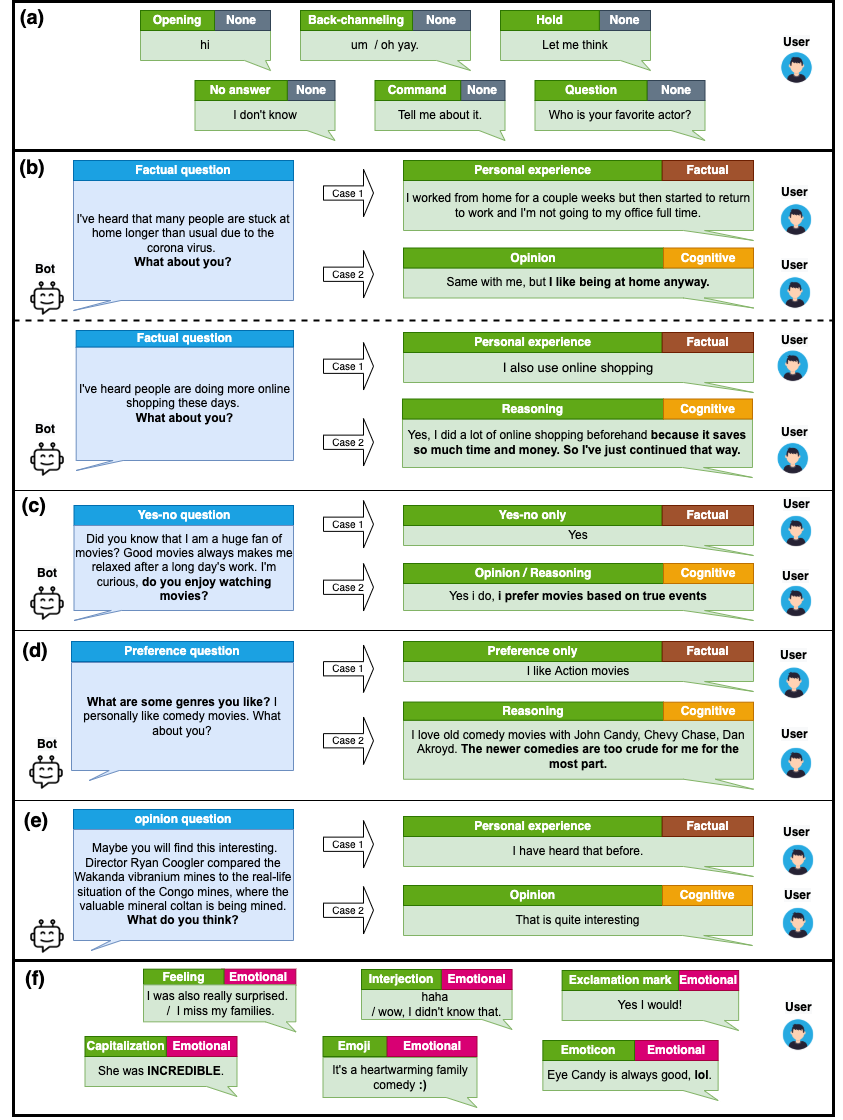}
    \caption{Users' self-disclosure level annotation scheme. \textbf{(a): None disclosure.} This includes opening, back-channeling, hold, no answer, command, and question. \textbf{(b - e): Factual and cognitive disclosure.} The two levels are dependent on the dialog context (i.e., the chatbot's response in the previous turn). In general, when users voluntarily provide opinion and reasoning, it's considered cognitive disclosure. If users' response is disclosing barely minimum information, it's consider factual even when answering preference or opinion questions. \textbf{(f) Emotional disclosure.} This includes feeling, interjection, exclamation mark, capitalization, emoji, and emoticon. 
    }
    \label{fig:annotation_scheme}
\end{figure*}

Following the three-level self-disclosure framework, we designed a single-label annotation scheme with four labels: \textit{none}, \textit{factual}, \textit{cognitive}, and \textit{emotional}. The self-disclosure labels for users align with the levels we designed for the chatbot to ensure consistency. 
Figure \ref{fig:annotation_scheme} shows the annotation schemes.  \textbf{``None''} self-disclosure indicates the user does not share anything about himself/herself, such as opening, back-channeling, hold, command, and question. \textbf{``Factual''} and \textbf{``cognitive''} levels depended on contextual information (i.e., the chatbot's response in the previous turn). Following the definition from SPT, self-disclosure is the ``intentional'' revealing of personal information. Therefore, in our annotation scheme, when users intentionally provide opinion, reasoning or preferences, it's considered cognitive disclosure. In contrast, if users' responses disclose barely minimum information, it's consider factual even when answering preference or opinion questions. For example, when the user only shared factual experience to a bot's question, answered yes or no to a yes/no question, or selected a preference without any explanation, it was annotated as ``factual''. As for \textbf{"emotional"} level, when participants' emotions (e.g., revelation of feelings, usage of exclamation marks, interjection, emoji, and emoticon) were contained in the message, it was considered ``emotional''.  

When annotating the data, since users' utterance at each turn may include multiple sentences with different self-disclosure levels, we first segmented it into sentences using NLTK sentence tokenizer, and then annotated each sentence segment of its self-disclosure level. 

\subsection{Dataset} 
We deployed a pilot study of 41 tasks (14  FD, 13 CD, and 14 ED) to collect training data. To ensure annotation reliability, two dialog experts (co-author of this paper) annotated 76 randomly selected sentence segments and reached a Cohen's kappa of 71.1\%, indicating substantial agreement. After the two annotators discussed annotation discrepancy and reached a consensus, one annotator annotated 535 more segments, resulting in a total of 611 annotated sentence segments (263 factual, 200 cognitive, 90 emotional, 58 none). We split the annotated samples into training/development set with a 75/25 ratio. Since the labels were highly imbalanced, we balanced the training data by oversampling minority classes to the same amount as the majority class, resulting in a total of 800 training examples (200 examples for each label). 





\subsection{Classifier}
To build a self-disclosure classifier, we started with a BERT-based neural model (\texttt{bert-base-cased}) pre-trained with Wikipedia and BookCorpus \cite{devlin2018bert}, and fine-tuned it with the 800 training examples for the classification task. The model used 12 layers with 12 attention heads and a hidden size of 768. The fully connected layers used a dropout rate of 0.1. 

As the self-disclosure levels were context-dependent, we included the bot's utterance of the last turn, and the previous user utterance segments of the current turn in the input to classify each user utterance segment. Inspired by \cite{yu2021midas}'s method of context representation, we appended the bot`s last utterance (\textit{bot\_last\_turn}), the user's utterance prior to the target segment of the same turn (\textit{user\_prev\_segs}), and the target user segmented text (\textit{user\_cur\_seg}) as \textit{[CLS] bot\_last\_turn : user\_prev\_segs [SEP] user\_cur\_seg [SEP]}. If there was no previous user segment, we then put an \texttt{EMPTY} token in the \textit{user\_prev\_segs}. After training, the model reached a macro average F1 score of 79.6\% (precision 78.8\%, recall 80.5\%). 
Considering some types of emotional self-disclosure were context-independent and can be easily distinguished, we patched the classifier with rules to enhance the performance. We used the \texttt{emot} \cite{emot} library to detect emoji, and regular expressions to detect exclamation mark and interjections such as \textit{ha, wow, lol}. This led to a slightly improved performance with a macro average F1 score of 81.7\% (precision 80.4, recall 83.2\%). 

\begin{figure}
    \centering
    \includegraphics[width=0.45\linewidth]{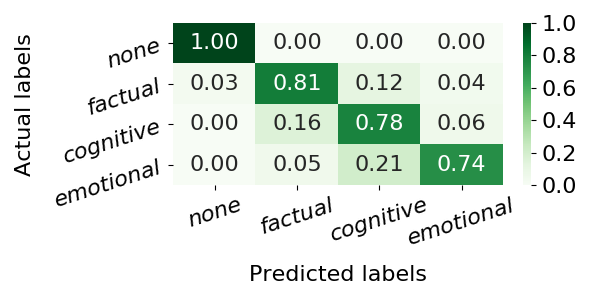}
    \caption{Confusion matrix of the self-disclosure classifier}
    \label{fig:confusion_matrix}
\end{figure}

The confusion matrix is shown in Figure \ref{fig:confusion_matrix}. We found that most of the misclassifications occurred between adjacent levels. For instance, sentences with ``emotional'' levels were sometimes classified as ``'cognitive'' but seldom classified as ``factual'', and the ones with ``cognitive'' levels were occasionally detected as ``factual'' but never detected as ``none''.  
This might be due to the ambiguity between adjacent levels. For example, in this case: \textit{Bot: Do you think your diet has changed since you've been staying at home a lot? User: Yes, less fast food.}, the user provided explanation on how his/her diet changes to a yes/no question, so it was labeled as ``cognitive'', but the model detected it as ``factual'' potentially because the explanation was also a fact.


In the dialog system, the classifier detected participants' self-disclosure level in real-time whenever the system received participant's utterance. If the participant utterance had multiple sentence segments in the same dialog turn, the classifier first detected self-disclosure level for each segment (segmented with NLTK sentence tokenizer) and then selected the highest self-disclosure level to represent the self-disclosure level for that turn.

\section{Effects of Self-disclosure}

\subsection{Self-disclosure Reciprocity (RQ1)}
To understand how participants reciprocate a chatbot's self-disclosure, we measured users' self-disclosure level and word length per dialog turn.
As described in \ref{sec:dialog_sessions}, there were some turns where the bot did not perform self-disclosure. 
To understand how users reciprocate, we combined all dialogs in FD, CD, ED and categorized the self-disclosure levels of users' turns by bot's self-disclosure level expressed every turn.



\begin{figure}[t]
    \centering
    \includegraphics[width=0.7\linewidth]{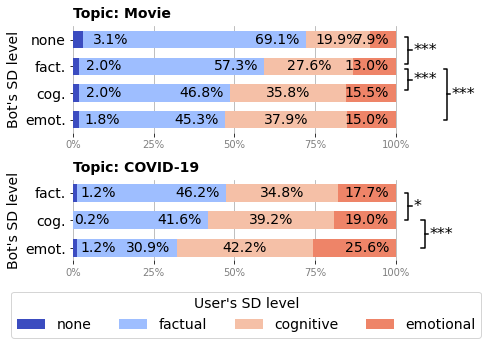}
    \caption{Participants' self-disclosure (SD) level distribution in reaction to a bot's SD at different levels.  ``fact.'' stands for factual disclosure, ``cog.'' stands for cognitive disclosure, and ``emot.'' stands for emotional disclosure. }
    \label{fig:reciprocity_classified}
\end{figure}

\begin{figure}[ht]%
\centering
\includegraphics[width=1.0\linewidth]{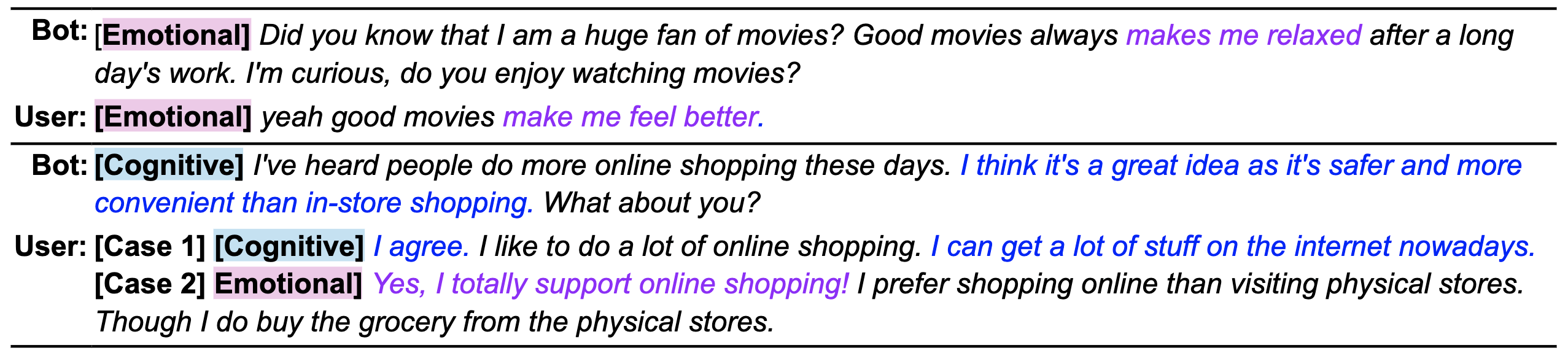}
    \caption{Dialog excerpts of how participants reciprocate a bot's cognitive and emotional self-disclosure.}
    \label{fig:dialog_excerpt}
\end{figure}

To evaluate how users reciprocate to chatbot's different self-disclosure levels, we first identified users' self-disclosure levels in each turn with a self-disclosure classifier (Section \ref{sec:self-disclosure-clssifier}). Then we performed nine 2-by-4 chi-square tests (six in Movie and three in COVID-19) to compare users' self-disclosure level distribution between every two combinations of the bot's level. We found that users significantly reciprocated to the bot's self-disclosure level 
(Figure \ref{fig:reciprocity_classified}). In movie dialogs, users' responses following bot's cognitive and emotional self-disclosure displayed higher ratio of cognitive and emotional levels than responses after bot's factual self-disclosure ($\chi^2(3, N=1855) = 21.63, p<.001$,  $\chi^2(3, N=1788) = 28.31, p <.001$). In COVID-19 dialogs, the reciprocity effects were more salient between adjacent levels. As the bot's self-disclosure levels increased, the users showed an increased likelihood of higher levels of self-disclosure. ($\chi^2 (3, N=1349) = 8.87, p < .05$ between cog. and fact., and $\chi^2(3, N=1205) =20.86, p < .001$ between emot. and cog.) This suggests that users reciprocate the bot's self-disclosure levels. Examples of how users reciprocated the bot's self-disclosure levels are presented in Figure \ref{fig:dialog_excerpt}. 

It should be noted that the dominant user disclosure level does not seem to always correspond to the bot's self-disclosure level. As the bot's disclosure level increase (e.g., from factual to cognitive; or from cognitive to emotional), the dominant level of users' self-disclosure is usually one or two levels lower than the bot's. That is, matching the bot's high level of self-disclosure may be difficult for users. Nevertheless, the bot's high self-disclosure level may still serve as a driving force to encourage users to disclose more. 
This may suggest that when the goal is to encourage people to disclose more, it may be more effective if the bot uses higher self-disclosure levels than simply matching the expected disclosure level at the user side.

\begin{table}[t]
\centering
\begin{tabular}{lrr}
\hline
\multirow{2}{*}{\textbf{Bot SD level}} & \multicolumn{2}{c}{\textbf{Word count}}                                     \\ \cline{2-3} 
                                       & \multicolumn{1}{l}{\textbf{Movies}} & \multicolumn{1}{l}{\textbf{COVID-19}} \\ \hline
\textbf{None}                          & 5.86                                & \multicolumn{1}{c}{na}                 \\ \hline
\textbf{Factual}                       & 6.71                                & 9.72                                  \\ \hline
\textbf{Cognitive}                     & 8.49                                & 11.48                                 \\ \hline
\textbf{Emotional}                     & 8.93                                & 12.36                                 \\ \hline
\end{tabular}

\caption{
Average user response length after a bot's self-disclosure at different levels.
}
\end{table}
\label{table:reciprocity_word_count}

Word count of user responses after bot's cognitive and emotional disclosure turns  were both significantly higher than FD's (p $<$ 0.01) in both topics, which aligned with previous results. Cognitive and emotional levels did not differ significantly from each other.

\subsection{Engagement and Perception of Bot (RQ2)}

\begin{figure*}[t]
     \centering
     \begin{subfigure}[b]{0.48\textwidth}
         \centering
         \includegraphics[width=\textwidth]{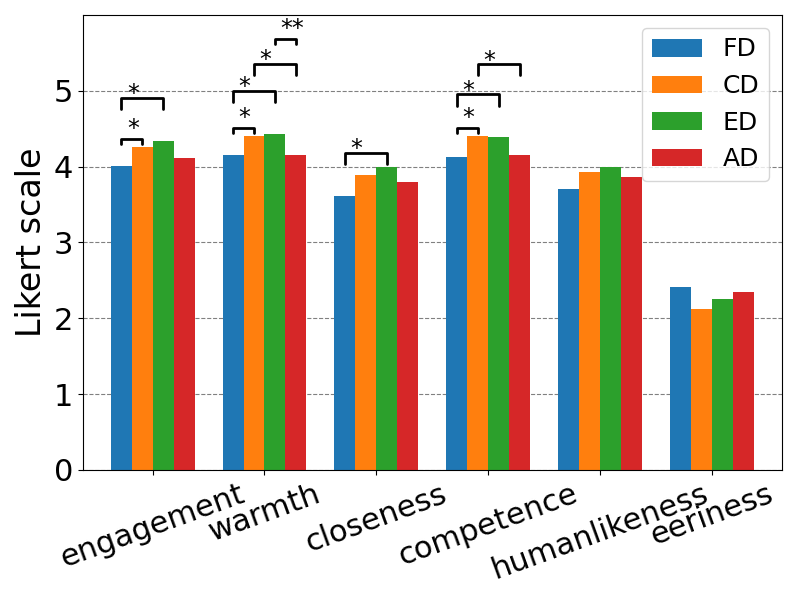}
         \caption{Movie}
         \label{fig:survey_result_movie}
     \end{subfigure}
     \hfill
     \begin{subfigure}[b]{0.48\textwidth}
         \centering
         \includegraphics[width=\textwidth]{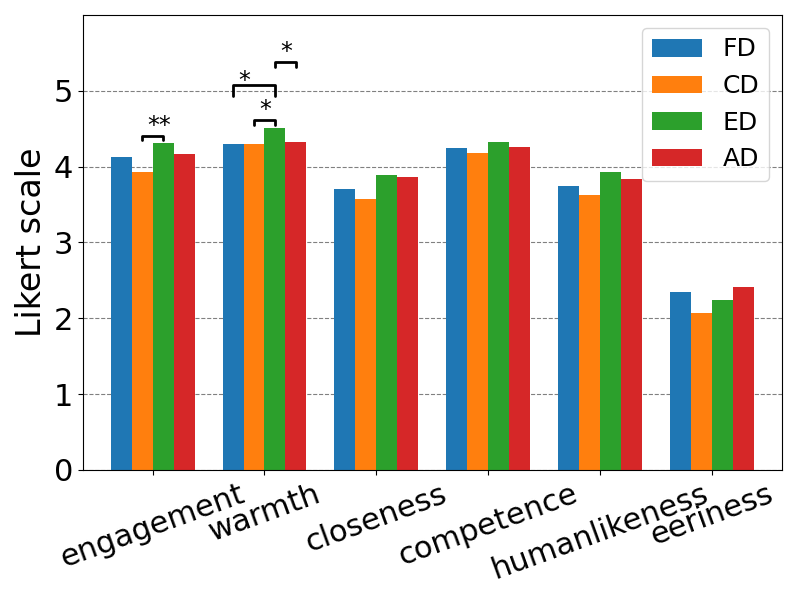}
         \caption{COVID-19}
         \label{fig:survey_result_covid}
     \end{subfigure}
        \caption{Results of conversation engagement and users' perception of the bot. The score of each construct is the average score of three measurement items with 5-point Likert scales. (a) Users' engagement and perception of the bot in movie dialogs. (b) Users' engagement and perception of the bot in COVID-19 dialogs. FD stands for factual disclosure, CD stands for cognitive disclosure, ED stands for emotional disclosure, and AD stands for adaptive disclosure. *: $p <.05$, **: $p < .01$, and ***: $p < .001 $.  }
        
        \label{fig:survey_result}
\end{figure*}

We performed two-tailed t-tests between different levels of the chatbot's self-disclosure and found that the chatbot's self-disclosure significantly affected users' engagement and perceived warmth (Figure \ref{fig:survey_result}). We also found significant interaction effects between topic and self-disclosure level for most of the constructs, so we evaluated the results of the two topics separately.

\subsubsection{Engagement}
In movie sessions (Figure \ref{fig:survey_result_movie}), posthoc analysis showed that both cognitive and emotional disclosures led to higher user engagement than factual self-disclosure ($t(195)=1.99, p<.05$ and $t(189)=2.51, p<.05$ respectively), while the user engagement did not differ between cognitive and emotional disclosure. In addition, user engagement increased when the bot's movie preference was similar to theirs. The following quotes from the participants illustrate how participants perceived the bot's disclosure of opinion, preference, and emotion. Overall, participants appeared to accept and appreciate that the chatbot was designed to exhibit cognitive self-disclosures:
\begin{quote}
``\textit{I liked how the Bot asked me questions about myself and also included its own opinions and views.}'' (P05, CD group, movie)
\end{quote}
Some participants even expressed negative perception toward situations where the chatbot didn't have cognitive self-disclose:
\begin{quote}
``\textit{I did not like that the chat bot did not have opinions of their own. }'' (P37, FD group, movie)
\end{quote}
In consistent with the patterns shown by the quantitative results, chatbot's emotional self-disclosure appeared to be in particular welcoming:
\begin{quote}
``\textit{(I liked) The part that the BOT told me comedy is its favorite genre. }'' (P93, ED group, movie)\\
``\textit{(I liked that) The bot seemed to have more feelings. }'' (P305, ED group, movie)
\end{quote}


In COVID-19 sessions, the bot with emotional self-disclosure was significantly more engaging than the one using cognitive self-disclosure (p $<$ 0.01). This might be because COVID-19 was an issue that people suffered from, so people may have expected the bot to show more feelings and empathy than when discussing movies. For instance, two participants commented: 
\begin{quote}
  ``\textit{I liked that the bot showed emotion \& feeling towards people dining out.} '' (P291, ED group, COVID-19) \\
  ``\textit{I liked when bot was recommending something based on what was relevant to the conversation. It felt like it was reaching out and giving good emotional feedback.} '' (P137, ED group, COVID-19)
\end{quote} 


\subsubsection{Perceived Bot Warmth} 
The effect of the bot's self-disclosure on user perception of the bot's warmth was significant. Post-hoc analyses showed that for both movie and COVID-19 sessions, the bot was perceived warmer in emotional disclosure ($t(189)=2.53, p<.05$ and $t(195)=2.25, p<.05$ respectively) than in factual. 
The effect of emotional self-disclosure was even more significant than cognitive disclosure in COVID-19 sessions ($t(174)=2.10, p<.05$). This may be because COVID-19 was related to people's immediate welfare, and the bot's emotion revealed its caring for people. As noted by a participant, the bot was more appreciated when it showed emotions:
\begin{quote}
``\textit{I loved when the bot empathized with being stuck at home. Very relatable. }'' (P240, ED group, COVID-19) \\
``\textit{I liked the part where I felt that even though it was a bot I was talking to, there was some empathy and compassion. }'' (P72, ED group, COVID-19) 
\end{quote}
Although there is no significant effect between cognitive and emotional self-disclosure in movie sessions, empathy is expected if the participants reveal negative feelings. As one stated about the part they dislike the most of the bot: 
\begin{quote}
``\textit{I think the bot did'nt show too much empathy towards my answers. }'' (P95, CD group, movie)
\end{quote}

Also, in movie sessions, the participants perceived the bot with cognitive disclosure warmer than the one with facts only. This could be because, in movie sessions, the bot revealed its own preference towards movies, actors, and genre, rather than simply providing facts. 

Moreover, the results showed that users in the adaptive disclosure condition perceived the bot as less warm than users in the emotional disclosure condition ($p < .01$ in movie, $p < .05$ in COVID-19). Especially in movie dialog, participants perceived the adaptive bot to be less warm than the cognitive bot ($p < .05$). This may be because the adaptive bot was more ``passive'' in disclosing itself than the cognitive and emotional bot. Instead of taking the initiative to disclose more about itself, the adaptive bot disclosed more only when the user did so. This may have led to a lower perception of its warmth. 
Therefore, we suggest that future dialog design starts with a higher self-disclosure to enhance users' experience.

\subsubsection{Perceived Closeness, Competence, Human-likeness, and Eeriness}
In movie dialogs, the perceived closeness between humans and the bot was significantly higher when the bot engaged in emotional self-disclosure than the bot's factual disclosure ($t(189)=2.49, p<.05$). We also found that people perceived the bot to be more competent in movie sessions when the bot showed cognitive or emotional self-disclosure than factual disclosure ($t(195)=2.58, p<.05$ and $t(189)=2.25, p<.05$ respectively). This means that sharing opinions or emotions made the conversation more coherent and reasonable. Although there was no significant effect of human-likeness in separate topics, we combined results from both sessions and found that the emotional bot was perceived more humanlike than the factual bot ($t(380)=2.09, p<.05$). There was no significant difference in perceived eeriness across conditions.

\subsubsection{Users' Responses to Chatbots' Self-disclosure}
In addition to the quantitative result, we also investigated the dialog between human-chatbot, and found users had interesting reactions when the chatbot revealed personal information that was out of users' expectations. For example, in the following conversation, the user was surprised when hearing the chatbot also ``got stuck at home'': 

\begin{quote}
\textit{\textbf{Chatbot:} ... I've heard that many people are stuck at home longer than usual due to the corona virus. What about you? \\
\textbf{User:} I was for a time \\
\textbf{Chatbot:} \underline{I was stuck at my house for a week.} I feel really sorry to see people forced to stay home, but I think it could be a good chance to watch more TV, and gain some new skills like cooking or gardening. I'm curious, what do you usually do to fill your time while staying at home? \\
\textbf{User}:  \underline{I didn't know chatbots had a house!} I just stayed home, worked on my computer, read books and let the dog in-out-in-out-in-out...
}
\end{quote}



\subsection{Recommendation Effectiveness (RQ3)}

\begin{figure}
    \centering
    \includegraphics[width=0.5\linewidth]{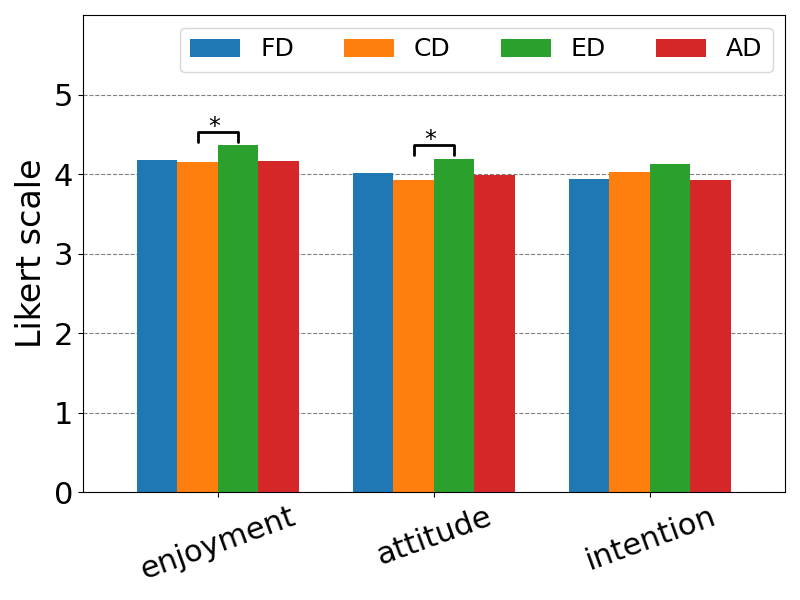}
    \caption{Overall recommendation effectiveness in both movie and COVID-19 tasks.  The score of each construct is the average score of three measurement items with 5-point Likert scales. Enjoyment stands for how users enjoys receiving the recommendation, attitudes stands for how useful the users consider the recommendation, and intention stands for how much users intend to follow the recommendation. FD stands for factual disclosure, CD stands for cognitive disclosure, ED stands for emotional disclosure, and AD stands for adaptive disclosure.  *: $p <.05$, **: $p < .01$, and ***: $p < .001 $.  }
    \label{fig:recommendation_effectiveness}
\end{figure}

We conducted multiple two-tailed t-tests between groups and found a significantly greater effect of bot's emotional self-disclosure on participants' enjoyment ($ t(350) = 2.01, p<.05$) and attitude ($t(350) = 2.42, p<.05$) toward the recommendation, compared to the effect of bot's cognitive self-disclosure (Figure \ref{fig:recommendation_effectiveness}). This means that people were more likely to enjoy and have more positive attitudes toward a bot's recommendations when a bot engaged in emotional disclosure than when the bot disclosed cognitive status. However, there was no significant difference in intention to follow recommendations. 

When answering ``Which part of the conversation did you like best? '', we found many participants expressed that the bot's recommendation was the part they like the most regardless of the bot's self-disclosure level. As some participants commented: 
\begin{quote}
\textit{"The recommendation was very pleasant. " (P120, FD group, COVID-19)} 

\textit{"(I liked) When the bot gave me advice to being happy and not letting quarantine get to me." (P304, FD group, COVID-19)} 
\end{quote}
Some recommendations from the chatbot were considered as being functional and helpful:
\begin{quote}
\textit{"I liked when the bot recommended me self-care methods. " (P2, FD group, COVID-19)} 

\textit{"I thought the part towards the end where the bot suggested I unplug from technology and go outside for at least 30 minutes a day was the best. " (P192, CD group, COVID-19)} 
\end{quote}
In the movie dialog, participants were quite open to recommendations of movies from the chatbot:
\begin{quote}
\textit{"I liked getting a movie recommendation at the end of the conversation. " (P251, FD group, movie)}

\textit{"(I liked) The recommendation. I will check out that movie. " (P370, ED group; movie)} 
\end{quote}

\section{Discussions}
Our results show that, compared to a chatbot that only reveals facts, opinions, or passively matches users' self-disclosure, a recommendation chatbot that consistently reveals emotions throughout the conversation leads to higher self-disclosure reciprocity, conversational enjoyment, and recommendation effectiveness. 
Our work departs from prior work in the following ways. First, we investigated the effects of a chatbot's self-disclosure by identifying three distinct levels of self-disclosure, as suggested by the Social Penetration Theory, instead of just one general category (i.e., with versus without self-disclosure) \cite{ravichander2018empirical} or two  (high-level versus low-level) \cite{lee2020hear}, and thus provided a more nuanced picture of how bots can leverage various self-disclosures to influence users. A notable addition was the adaptive self-disclosure bot which dynamically adapted the self-disclosure strategy to participants' levels of self-disclosure. Also, users' reciprocal behaviors were coded using a reliable annotation scheme for disclosure to examine how users adapt their disclosure to the bot's disclosure levels. Lastly, we examined how people reacted to the bot's self-disclosure and how different levels of bot's self-disclosure changed people's perceptions of the bot and its recommendation effectiveness.

\subsection{Effects of Self-disclosure on Reciprocity (RQ1)}
With regard to RQ1, we found that as the bot's self-disclosure level increased in depth, participants were more likely to reciprocate with an increased level of self-disclosure. Our findings align with previous research that demonstrated that chatbot's self-disclosure can be reciprocated by users \cite{lee2020hear, ravichander2018empirical}. More specifically, as shown in \cite{lee2020hear}, deeper level self-disclosure by the bot can facilitate deeper level self-disclosure from users. Such a finding implies that chatbots may strategically start with deeper levels of self-disclosure (e.g., emotional disclosure) to motivate deeper self-disclosure from users. However, it should be noted that the extent to which users reciprocate to a bot's self-disclosure level may differ by conversational topics. We found that the users' reciprocal behaviors were more salient in COVID-19 related conversations than in movie-related conversations. Thus, investigating the local effects specific to different conversational topics and tasks would be necessary in future studies.

\subsection{Effects of Self-disclosure on Engagement and Perception of the Chatbot (RQ2)}
With regard to RQ2, on how chatbot's different levels of self-disclosure influence people's engagement with and perception towards a bot, our results highlight the effect of bot's emotional self-disclosure. More specifically, we found that user engagement was the highest when they interacted with the emotional disclosure bot. In addition, users perceived the emotional self-disclosure bot to be warmer than factual disclosure in both conversational tasks. Such results are aligned with previous research \cite{ho2018psychological}, which indicated that participants' perception of chatbot's warmth was significantly higher after emotional disclosure compared to factual disclosure. This can be explained by Computers as Social Actors (CASA) paradigm which argues that humans perceive and interact with computers just as humans would interact with other humans \cite{nass1994computers, reeves1996media}. This framework suggests that disclosure processes and outcomes would be similar regardless of whether the partner is a human or a bot \cite{ho2018psychological}. Given that emotional self-disclosure is at the core of intimate relationship development and more effective to cultivate strong connection between people \cite{kim62song}, such significant effect of emotional disclosure on interpersonal outcomes (i.e., engagement and warmth) may have emerged in human-chatbot conversations as well.

Moreover, we found that the effects of different levels of self-disclosure might be different between various types of topics. For example, cognitive disclosure also leads to better conversational enjoyment than factual disclosure when discussing digital entertainment such as movies. However, this is not the case when discussing daily life or well-being topics such as Covid-19. Although the difference is not statistically significant, users perceived the conversation as less enjoyable when the bot performed cognitive disclosure without emotional acts in our study. This might be because people could have expected to engage in the conversation to acquire information about movies and be prepared to enjoy such conversations as the topic was about entertainment. Whereas in daily life topics like lives during the Covid-19 pandemic, since chatbots are virtually not human and cannot experience daily activities as people, people might consider chatbots being unrelatable to their experiences. Hence, chatbots' opinions may not help improve conversational engagement and perceived warmth in this context. 

However, when a chatbot revealed emotions regarding daily life events, people started to perceive the chatbot warmer, and their engagement in conversation increased significantly. As shared by some participants, they liked how the emotional bot showed empathy and compassion even if they knew they were talking to a chatbot. 
Similarly, humans' understanding of chatbots' identity might influence the extent to which people feel connected with the chatbot based on the genre of conversation. Our study found that emotional disclosure by chatbot improves human-chatbot closeness in digital entertainment conversation more evidently than in daily life conversation.

\subsection{Effects of self-disclosure on Recommendation Effectiveness (RQ3) } 
The significant effects of bot's emotional self-disclosure were persistent on recommendation effectiveness as well (RQ3). We found that users enjoyed listening to chatbot's recommendation when the chatbot shared its emotions (i.e., emotional disclosure) compared to when the chatbot shared its thoughts (i.e., cognitive disclosure). More importantly, our result showed that participants had more positive attitudes towards the chatbot's recommendation when the bot disclosed emotions than thoughts. Prior work \cite{hayati2020inspired} on human-human recommendation dialog has shown that strategies involving self-disclosure have positive effect on users' acceptance of the recommendation. Our study showed that chatbot's self-disclosure has similar effects on users' acceptance of the recommendation as human's self-disclosure, which aligns with the CASA paradigm \cite{ho2018psychological}. In addition, given that the chabot's emotional self-disclosure induced a more positive attitude towards the recommendation than cognitive disclosure, we speculate that a deeper level of self-disclosure may have enhanced users' positive perception towards the bot, which in turn increased the bot's persuasiveness. Lastly, with regard to users' intention to follow the recommendation. there was no significant difference between different levels of self-disclosure.
This might be because the recommendation is provided voluntarily to the users instead of being requested by the users intentionally.  Despite this, there is still a trend of increasing intention as the self-disclosure levels arise. 



\subsection{Design Implications}
\paragraph{Proactive verses Matching self-disclosure level from Chatbots}
Our result has already indicated that human tends to reciprocate with the chatbot's self-disclosure level. That is, when chatbots take the initiative to disclose with higher levels (such as the CD and ED chatbot), people reciprocate with higher self-disclosure. On the other hand, the reciprocity of the AD chatbot works in the reverse direction. The AD chatbot waits for the users to take the initiative to raise the disclosure to a higher level. And when the users do so, the chatbot reciprocates with matching levels. Our results show that the AD bot leads to less engagement and warmth than the CD and ED bot. 
This is consistent with our results that a chatbot consistently revealing its emotion throughout the interaction (ED bot) is more effective than the ones with only factual or cognitive disclosure. The reason might have been due to that the ED chatbot is proactive in disclosing more, while the AD bot is comparatively passive in disclosing itself. The result also shows that the performance of the AD bot is equivalent to or slightly better than the FD bot. This might be because the FD bot is unwilling to disclose more regardless of how the users disclose, while the AD bot still follows the user's disclosure level and may have been considered more responsive.




\paragraph{Leveraging Real-time Classifiers during Interaction}
It's also worth noting that in our work, we implemented a  self-disclosure classifier, and incorporated it into our chatbot system to identify the users' self-disclosure level in real-time during the conversational social interaction, and adapt the disclosure level from chatbot accordingly in one of the experimental conditions. By doing so, we demonstrated the technical possibility to monitor the user's status in real-time with a machine learning model, and adapt the chatbot system accordingly. Although the adaptive chatbot is technically more complicated than other chatbots, it does not lead to better performance in terms of recommendation. This might have confirmed that, in conversational recommendation, a technically more sophisticated system is not necessarily socially more beneficial or feasible. Alternatively, with real-time self-disclosure identification, we could potentially restrict chatbot's intervention as well in scenarios and applications where the goal is to maintain the neutrality of conversation.


\paragraph{Chatbots' Identity: What Personal Information can Chatbots' disclose?}
Our work also revealed that users have assumptions of a chatbot's identity and their understanding of what a chatbot can or cannot do. For instance, one user got surprised when hearing a chatbot say it got stuck at its house, as people assumed a chatbot does not have a house. This suggests that designing chatbots' self-disclosure is a non-trivial task. Future research could explore what chatbot behaviors are considered reasonable from a human perspective, and the effect of anticipated and non-anticipated self-disclosure on the engagement, perception, and recommendation effectiveness.  

\subsection{Ethical Considerations, Limitation and Future Work}
Given the fast development of automated chatbot systems that can influence human beliefs and actions, an ethical design principle must be in place throughout all stages of the development and evaluation. Key ethical considerations include having transparency and user trust, ensuring user safety, and minimizing biases \cite{zhang2020artificial,oh2021systematic}. In designing the movie and COVID-19 recommendation chatbots, our goal was to help users relieve their stress and anxiety during the pandemic. To ensure transparency and gain user trust, participants were fully informed of the purpose of the study and the intent of the chatbot. To ensure user safety, continuously monitored the conversations to make sure the conversations complied with recommended ethical standards and did not result in unintended harm. Lastly, we ensured the bot responses were appropriate and nondiscriminative to minimize potential bias.

We acknowledge limitations of our current approach. First, in our work, as the self-disclosure levels are incremental between conditions due to the consideration of naturalness, the response length of a chatbot with higher disclosure level is averagely longer than the chatbots with lower levels. Future work could consider the influence from the chatbot's response length in their experimental design. Second, we include two intrinsically different topics, movies and COVID-19, to demonstrate the generalizability of the effects of different self-disclosure levels, and our results suggest that emotional disclosure is the most effective element. To gain more insights into how different types of emotions influence different topic categories, future work could explore if a chatbot's positive and negative emotion impact the effect of recommendation differently. For example, in the application of mental health chatbots, users might have more negative feelings, and a chatbot's types of emotions might impact users' acceptance to the chatbot's suggestions.




\section{Conclusion} 
This study experimentally tested how people reacted to chatbot's different levels of self-disclosure with a text-based sociable recommendation chatbot. We showed that peoples' self-disclosure levels were positively correlated to the chatbot's self-disclosure levels. Also, we found that higher levels of self-disclosure led to more engaging conversations and warmer bot perception. Lastly, emotional self-disclosure significantly enhanced people's enjoyment and attitude to a chatbot's recommendations. We believe our work provides a better understanding of how a bot's self-disclosure can be leveraged to encourage users' self-disclosure, improve users' perception of a chatbot, and enhance the persuasiveness of the recommendation.


\bibliographystyle{ACM-Reference-Format}
\bibliography{main}

\clearpage
\appendix

\section{Appendix}

\subsection{Dialog Design}
\label{sec:full_dialog_design}
\begin{figure*}[ht!]
    \centering
    \includegraphics[width=\linewidth]{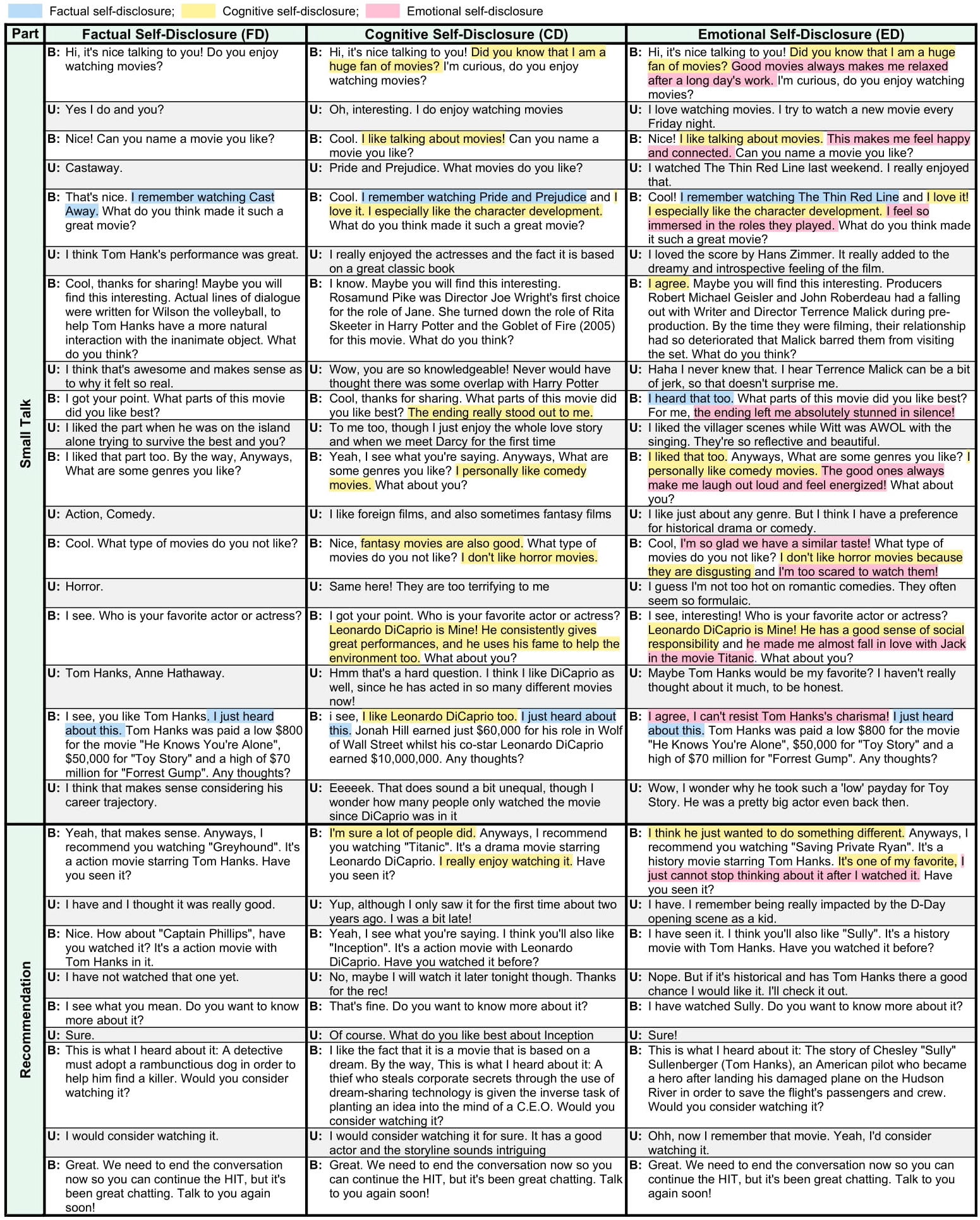}
    \caption{Movie Dialog}
    \label{fig:dialog_example_movie}
\end{figure*}

\begin{figure*}[ht!]
    \centering
    \includegraphics[width=\linewidth]{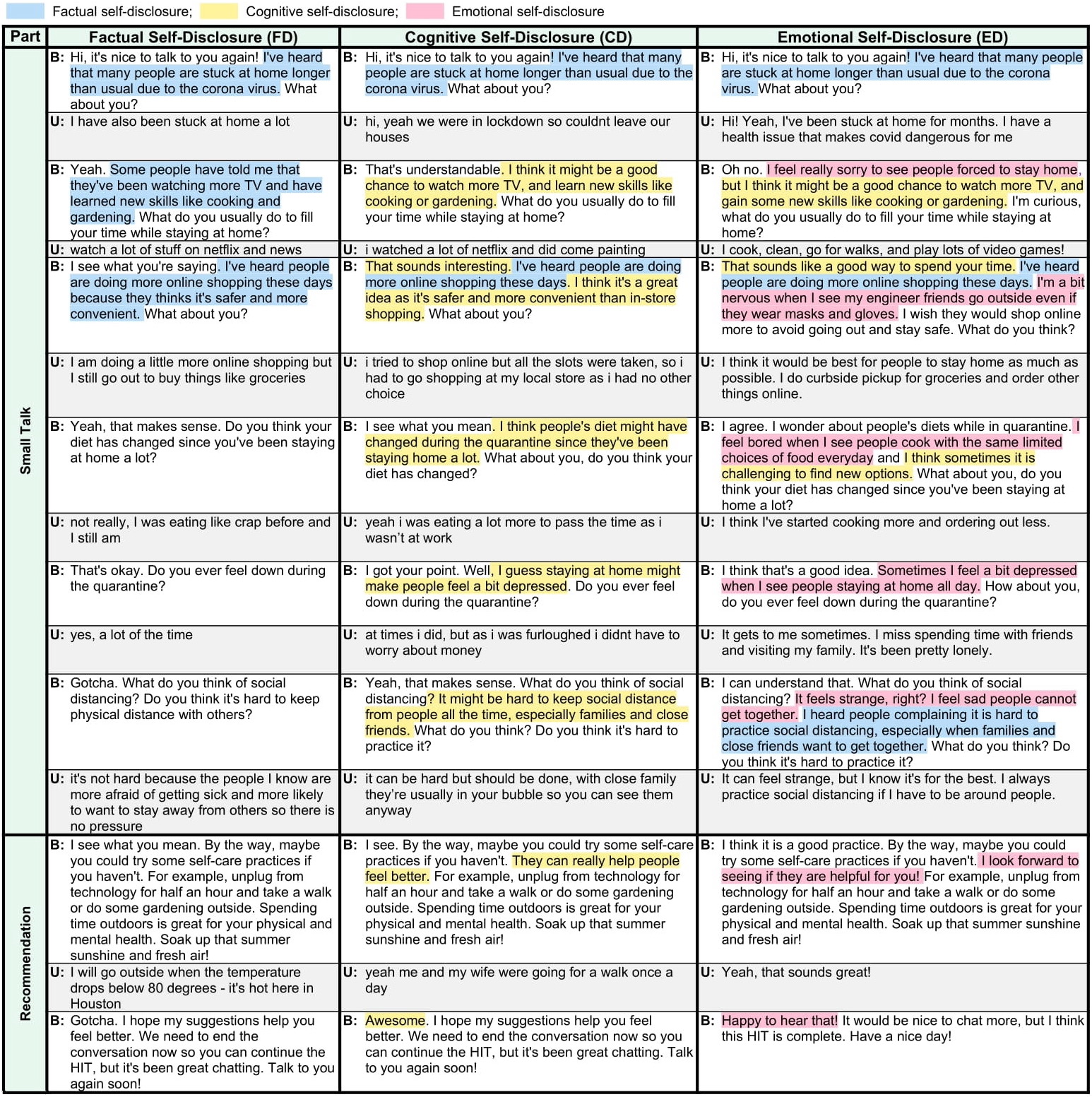}
    \caption{Covid Dialog}
    \label{fig:dialog_example_covid}
\end{figure*}

Example of full conversations on movie and Covid-19 can be found in Figure \ref{fig:dialog_example_movie} and Figure \ref{fig:dialog_example_covid} respectively.

\subsection{Recommendation}
\label{sec:movie_recommendation_rule}
The bot gave recommendations at the end of both movie and COVID-19 sessions.
In movie sessions, the bot recommended a movie to the participant and provide a movie summary if the participant is interested. Dialog example was demonstrated at the end of  Figure \ref{fig:dialog_example_movie}. The recommended movie was selected based on the following rules. If the participants' have a favorite actor, the bot first recommends the most popular movie the actor has acted. Otherwise, the bot recommends the most popular movie in the participants' favorite genre. If the participants' do not have a favorite actor or genre, the bot recommends movies based on the previously discussed movie.  If the participants have watched the recommended movie, the bot will recommends another one. We use TMDB's \texttt{discover} and \texttt{movies} APIs for the recommendations.

In COVID-19 sessions, self-care practices was suggested as shown at the end of Figure \ref{fig:dialog_example_covid}.

\subsection{Acknowledgement Generator}
\label{subsubsec:acknowledgement_generator}

To acknowledge participants' general utterances that were not questions, a generation model, Blender \cite{roller2020recipes} , was used in the first place to generate more engaging responses ((See \ref{subsubsec:blender_model}) for more details). If no response was generated, we then used hand-written acknowledgment templates designed either specifically for each dialog state or very common participant intents. For example, if the bot asked the participant about his/her favorite actor and the participant answers ``Jennifer Aniston'', the bot would answer ``I see, I like Jennifer Aniston too.'' If the participant said ``I don't know'', which is a common intent, the bot would generate ``That's okay''. If the participant said, ``That's interesting!'', the bot would said ``I'm glad you like it. '' For participants utterances that were not common intents, the bot replied with general acknowledgment such as ``I see.'', ``Gotcha.''

\subsection{Question Handler}
\label{subsubsec:question_handler}
When handling participant's questions, we used backstory-database \cite{liang2020gunrock} and Amazon EVI \footnote{https://www.evi.com/} to handle questions associated with the bot's persona and factual questions, respectively. If no answer is retrieved, we then used a text generation model, Blender \cite{roller2020recipes}, to generate answers (See \ref{subsubsec:blender_model} for more details). By leveraging the model, the bot could handle a wider range of participant input and generated diverse responses. If no response was generated, the bot generated a rephrased answer to express that the bot does not have an answer. For example, if the participant asked, ``How long will the coronavirus last?'', the bot would answer ``I don't know how long will the coronavirus last.''). If the bot failed to generate a rephrased answer, it replied with a general answer such as ``Sorry, but I don't know much about that. ''

\subsection{Blender Generation Model}
\label{subsubsec:blender_model}
 The Blender model is an encoder-decoder model with 2.7B parameters trained on large amounts of human-human conversation data, and is adjusted to learn several conversational skills such as the ability to assume a persona, being knowledgeable and show empathy. We adjust the model following \cite{liang2020gunrock}'s method, so that it generates response with decent length and short latency without proposing new topic or asking questions. It also contains a rule-based filter to clean utterances that demonstrate a personality that clearly contradicts the personality of a chatbot, such as  ``I like to eat at home with my wife''. 
 In addition, to ensure the generated responses align with the experimental condition of self-disclosure level, we examined about 200 generated responses from the pilot study, and created a self-disclosure level filter to exclude responses with unmatched self-disclosure level for each experimental condition. As we found the generative responses revealing explicit self-disclosure levels follow specific patterns, we implement the filter with regular expressions to detect relevant keywords. For example, in the factual condition, we filter out sentences containing keywords of cognitive disclosures (e.g., agree, think, thought, like, love, prefer) and emotional disclosures (e.g., feel, sad, happy, glad, sorry).
 
\subsection{Movie Name Grounding}
\label{sec:movie_grounding}
Since the discussion about a movie takes half of the turns, to ensure the conversation is not too short, if the bot fails to recognize the mentioned movie, it would ask users to name another movie until it recognizes it and starts discussing it.




\end{document}